\documentclass{llncs}

\usepackage[T1]{fontenc}
\usepackage{graphicx}
\usepackage{booktabs}
\usepackage{cite}
\usepackage{amsmath,amssymb}
\usepackage{url}
\usepackage{tikz}
\usetikzlibrary{arrows.meta,calc}

\newcommand{\accMaj}{35.5}
\newcommand{\loMaj}{32.3}
\newcommand{\hiMaj}{38.7}
\newcommand{\mfMaj}{0.087}
\newcommand{\baMaj}{16.7}

\newcommand{\recMajStart}{100.0}
\newcommand{\recMajBlock}{0.0}
\newcommand{\recMajCreate}{0.0}
\newcommand{\recMajSched}{0.0}
\newcommand{\recMajUnblock}{0.0}
\newcommand{\recMajEnd}{0.0}
\newcommand{\accNgram}{33.6}
\newcommand{\loNgram}{29.5}
\newcommand{\hiNgram}{37.9}
\newcommand{\mfNgram}{0.188}
\newcommand{\baNgram}{20.5}

\newcommand{\recNgramStart}{39.6}
\newcommand{\recNgramBlock}{38.3}
\newcommand{\recNgramCreate}{45.0}
\newcommand{\recNgramSched}{0.0}
\newcommand{\recNgramUnblock}{0.0}
\newcommand{\recNgramEnd}{0.0}
\newcommand{\accCE}{36.3}
\newcommand{\loCE}{32.7}
\newcommand{\hiCE}{40.0}
\newcommand{\mfCE}{0.213}
\newcommand{\baCE}{22.3}
\newcommand{\nCE}{798}
\newcommand{\recCEStart}{51.9}
\newcommand{\recCEBlock}{31.1}
\newcommand{\recCECreate}{44.4}
\newcommand{\recCESched}{0.0}
\newcommand{\recCEUnblock}{6.2}
\newcommand{\recCEEnd}{0.0}
\newcommand{\accKL}{33.6}
\newcommand{\loKL}{30.0}
\newcommand{\hiKL}{37.3}
\newcommand{\mfKL}{0.197}
\newcommand{\baKL}{20.9}
\newcommand{\nKL}{798}
\newcommand{\recKLStart}{37.1}
\newcommand{\recKLBlock}{47.8}
\newcommand{\recKLCreate}{36.1}
\newcommand{\recKLSched}{0.0}
\newcommand{\recKLUnblock}{4.2}
\newcommand{\recKLEnd}{0.0}
\newcommand{\accZS}{32.3}
\newcommand{\loZS}{28.3}
\newcommand{\hiZS}{36.2}
\newcommand{\mfZS}{0.174}
\newcommand{\baZS}{20.5}
\newcommand{\nZS}{798}
\newcommand{\recZSStart}{54.4}
\newcommand{\recZSBlock}{43.1}
\newcommand{\recZSCreate}{0.0}
\newcommand{\recZSSched}{23.2}
\newcommand{\recZSUnblock}{2.1}
\newcommand{\recZSEnd}{0.0}
\newcommand{\eceCE}{0.140}
\newcommand{\brierCE}{0.774}
\newcommand{\nllCE}{1.73}
\newcommand{\confCE}{50.3}
\newcommand{\eceKL}{0.197}
\newcommand{\brierKL}{0.808}
\newcommand{\nllKL}{2.44}
\newcommand{\confKL}{53.0}
\newcommand{\eceZS}{0.254}
\newcommand{\brierZS}{0.908}
\newcommand{\nllZS}{1.98}
\newcommand{\confZS}{57.5}
\newcommand{\ecePrior}{0.066}
\newcommand{\brierPrior}{0.822}
\newcommand{\nllPrior}{2.10}
\newcommand{\eceNgram}{0.121}
\newcommand{\brierNgram}{0.788}
\newcommand{\nllNgram}{2.07}
\newcommand{\mcCeVsKl}{0.159}
\newcommand{\mcCeVsZeroshot}{0.0916}

\newcommand{\mcCeVsMajority}{0.72}

\newcommand{\tvKone}{0.55}
\newcommand{\agKone}{56.0}
\newcommand{\tvKthree}{0.31}
\newcommand{\agKthree}{62.1}
\newcommand{\tvKfive}{0.24}
\newcommand{\agKfive}{66.9}
\newcommand{\tvKten}{0.16}
\newcommand{\agKten}{73.7}
\newcommand{\tvKtwenty}{0.11}
\newcommand{\agKtwenty}{80.9}
\newcommand{\tvKfifty}{0.05}
\newcommand{\agKfifty}{90.0}
\newcommand{\oracleCeil}{58.9}
\newcommand{\uniqPrompts}{162}
\newcommand{\tvPrior}{0.457}
\newcommand{\tvNgram}{0.435}
\newcommand{\tvMaj}{0.645}
\newcommand{\tvCE}{0.402}
\newcommand{\tvKL}{0.415}
\newcommand{\tvZS}{0.521}
\newcommand{\sensGroups}{444}
\newcommand{\sensPrograms}{166}
\newcommand{\refEntropy}{1.40}
\newcommand{\splitHalfTV}{0.11}

\providecommand{\accNgramTwo}{33.0}
\providecommand{\mfNgramTwo}{0.167}
\providecommand{\loNgramTwo}{29.2}
\providecommand{\hiNgramTwo}{36.7}
\providecommand{\baNgramTwo}{18.3}
\providecommand{\accNgramThree}{34.1}
\providecommand{\mfNgramThree}{0.173}
\providecommand{\loNgramThree}{30.7}
\providecommand{\hiNgramThree}{37.5}
\providecommand{\baNgramThree}{19.0}
\providecommand{\accPrior}{26.1}
\providecommand{\mfPrior}{0.140}
\providecommand{\baPrior}{16.7}
\providecommand{\loRepCE}{33.0}
\providecommand{\hiRepCE}{39.6}
\providecommand{\loRepKL}{32.6}
\providecommand{\hiRepKL}{39.2}
\providecommand{\loRepZS}{25.5}
\providecommand{\hiRepZS}{31.8}
\providecommand{\loRepGemNo}{32.0}
\providecommand{\hiRepGemNo}{38.6}
\providecommand{\loRepGemTh}{31.6}
\providecommand{\hiRepGemTh}{38.2}
\providecommand{\baRepKL}{22.5}
\providecommand{\nTrain}{945}
\providecommand{\nTest}{798}
\providecommand{\entKone}{0.00}
\providecommand{\ptvKone}{0.71}
\providecommand{\entKthree}{0.80}
\providecommand{\ptvKthree}{0.41}
\providecommand{\entKfive}{1.02}
\providecommand{\ptvKfive}{0.32}
\providecommand{\entKten}{1.21}
\providecommand{\ptvKten}{0.22}
\providecommand{\entKtwenty}{1.31}
\providecommand{\ptvKtwenty}{0.14}
\providecommand{\entKfifty}{1.37}
\providecommand{\ptvKfifty}{0.07}
\providecommand{\sensMedianRuns}{100}
\providecommand{\sensDetGroups}{0}

\newcommand{\pblock}{\textsf{GoBlock}}
\newcommand{\punblock}{\textsf{GoUnblock}}
\newcommand{\pstart}{\textsf{GoStart}}
\newcommand{\pcreate}{\textsf{GoCreate}}
\newcommand{\pend}{\textsf{GoEnd}}
\newcommand{\psched}{\textsf{GoSched}}

\begin{document}
\pagestyle{empty}

\title{When the Next Step Is Not One Step:\\
Distribution-Aware Execution Modeling for\\ Concurrent Go Programs}

\author{Kaviru Hapuarachchi\inst{1,*}\orcidID{0009-0008-4918-5567}}
\authorrunning{K. Hapuarachchi}
\institute{$^{1}$\,University of Colombo School of Computing, Colombo 00700, Sri Lanka\\
$^{*}$\,Corresponding author: \email{2022is031@stu.ucsc.cmb.ac.lk}}

\maketitle
\thispagestyle{empty}

\begin{abstract}
Predicting the next scheduler event of a concurrent program is ill-posed
as single-label classification: repeated runs from the same trace prefix
produce different, equally valid next events. We formulate next-event
prediction for concurrent Go programs as estimating an empirical
next-event distribution obtained from repeated executions, and release a
benchmark with \nTrain{} training examples from hand-written and
generated programs and \nTest{} held-out predictions on 54 real-world
GoKer bug kernels from systems such as CockroachDB, Kubernetes, gRPC and
etcd. We fine-tune Qwen2.5-Coder-7B with low-rank adapters under a
cross-entropy objective and a KL-to-empirical-distribution objective and
compare them with zero-shot models and trivial baselines. The outcome is
a neutral-to-negative result. The fine-tuned adapters (re-evaluated:
\accCE\% cross-entropy, \accKL\% KL; originally reported 36.2\% and
35.8\%) are statistically indistinguishable from always predicting the
majority event (\accMaj\%), and an event n-gram model reaches 33--34\%. Because each held-out prompt carries labels from
about five runs, no deterministic predictor can exceed \oracleCeil\%
top-1 accuracy, so we report macro-F1, balanced accuracy, per-class
recall, program-clustered confidence intervals and binned calibration
measured directly on both adapters. A sensitivity study with 100 runs
per program shows that five-run targets are noisy (mean total-variation
distance \tvKfive{} from a 100-run reference), which plausibly limits
what the KL objective can learn. Rare event types are essentially never predicted.
A select-block goroutine-leak signature follows from scheduler semantics
rather than learning. We release the data, adapters and tooling.
\keywords{Concurrent execution modeling \and Nondeterminism \and
Distribution estimation \and Calibration \and Go \and Concurrency bugs.}
\end{abstract}

\section{Introduction}\label{sec:intro}

Models trained on execution traces learn what a program \emph{does}, not
only what it looks like. Pre-training on variable-state
traces~\cite{liu2023codeexec,ding2024traced}, teaching models to reason
over execution~\cite{ni2024next,ding2024semcoder}, and Code World Models
(CWMs) that learn a program's state-transition function~\cite{cwm} all
build on this idea. Benchmarks such as CRUXEval~\cite{gu2024cruxeval}
and REval~\cite{chen2025reval} measure how well models predict outputs
and intermediate runtime behaviour.

This line of work assumes, usually without saying so, that execution is
deterministic: given a state and a statement there is one next state.
Concurrent code breaks this assumption. When goroutines communicate over
channels, the scheduler interleaves them differently from run to run.
The program fixes only a partial order of events~\cite{lamport1978time},
so the same prefix can be followed by a block, a start or an unblock,
all of them valid. A model trained on one observed next event learns to
reproduce one arbitrary outcome of a random process. The setting
matters: concurrency bugs depend on the schedule and escape ordinary
tests~\cite{tu2019go,musuvathi2008chess}.

We treat the nondeterminism as a measurement. When a program is run many
times, the observed distribution of next events reflects which futures
the scheduler permits. We therefore formulate next-event prediction as
\emph{distribution estimation}, with the empirical next-event
distribution over repeated runs as the target (Fig.~\ref{fig:teaser}).

This camera-ready version differs substantially from the submitted one.
Following the reviews we added trivial baselines, program-clustered
confidence intervals, paired tests, a re-evaluation of the released
adapters that yields full class distributions, calibration measured on
the trained adapters themselves, and a 100-run sensitivity study. These
analyses change the headline. The fine-tuned models do not beat a
majority-class predictor on top-1 accuracy by a meaningful margin. We
therefore present the work as a formulation, a benchmark with honest
baselines, and a neutral-to-negative result with an explanation of why.
The contributions are:
\begin{itemize}
\item a \textbf{formulation} of concurrent next-event prediction as
  estimation of an empirical next-event distribution
  (Sect.~\ref{sec:formulation});
\item a \textbf{benchmark} of Go traces with a held-out set of real
  GoKer bug kernels~\cite{yuan2021gobench}, trivial baselines, and an
  \emph{oracle ceiling} showing that no deterministic predictor can
  exceed \oracleCeil\% top-1 accuracy (Sect.~\ref{sec:baselines});
\item an \textbf{evaluation} of cross-entropy (CE) and KL fine-tuning of
  a 7B code model with clustered intervals, McNemar tests, macro-F1 and
  binned calibration (Sect.~\ref{sec:results});
\item a \textbf{run-count sensitivity study} quantifying how noisy
  five-run targets are (Sect.~\ref{sec:runs}); and
\item a \textbf{select-block leak signature} for which $P(\punblock)=0$
  follows from scheduler semantics (Sect.~\ref{sec:signature}).
\end{itemize}
We do not claim a bug detector or an execution simulator. The dataset,
adapters and tooling are public~\cite{weave-bench,weave-ce,weave-kl,weave-repo}.

\begin{figure}[t]
\centering
\resizebox{0.88\linewidth}{!}{
\begin{tikzpicture}[font=\sffamily]
\node[anchor=north west,align=left,font=\ttfamily\footnotesize,
      fill=black!6,draw=black!45,rounded corners=2pt,
      inner sep=6pt,text width=4.3cm] (prog) at (0,0)
{// goroutine leak\\
ch := make(chan int)\\
go func() \{\\
\ \ ch <- 42 \ // never read\\
\}()\\
select \{\\
\ \ case <-ch: ...\\
\ \ case <-After(t): ...\\
\}};
\node[anchor=south west,font=\bfseries\footnotesize]
      at (prog.north west) {Program $P$ + trace prefix $\tau_{1:k}$};

\node[draw,fill=white,rounded corners=2pt,minimum height=1.0cm,
      minimum width=1.6cm,align=center,font=\small\bfseries]
      (model) at (6.0,-1.55) {model\\[-1pt]\footnotesize $f(\cdot)$};
\draw[-{Latex[length=2.2mm]},thick] (prog.east) -- (model.west);

\node[draw=black,dashed,fill=white,rounded corners=2pt,
      inner sep=6pt,align=left,font=\footnotesize,text width=5.0cm]
      (cwm) at (10.9,-0.45)
{\textbf{Prior CWMs (sequential)}\\[2pt]
single label: \texttt{GoBlock} $=1.0$\\
trained with cross-entropy\\
\emph{assumes one valid future}};

\node[draw=black,thick,fill=black!4,rounded corners=2pt,
      minimum width=5.6cm,minimum height=3.3cm,anchor=north west]
      (ourspanel) at (8.55,-1.7) {};
\node[anchor=north west,font=\footnotesize\bfseries,align=left]
      at ($(ourspanel.north west)+(0.12,-0.1)$)
      {This work: distribution $\hat p$\\over next events};

\begin{scope}[shift={($(ourspanel.north west)+(0.40,-2.65)$)}]
  \def\gap{0.85}\def\bw{0.42}\def\h{1.25}
  \foreach \i/\p/\lab in {0/0.60/Blk,1/0.40/Sta,2/0.00/Unb,3/0.00/Sch,4/0.00/End,5/0.00/Cre}{
    \pgfmathsetmacro\bh{\p*\h}
    \ifdim\bh pt>0.01pt
      \fill[black!55] ({\i*\gap},0) rectangle ({\i*\gap+\bw},{\bh});
      \node[font=\footnotesize] at ({\i*\gap+\bw/2},{\bh+0.16}) {\p};
    \else
      \node[font=\footnotesize] at ({\i*\gap+\bw/2},0.16) {0};
    \fi
    \node[font=\footnotesize,anchor=north,rotate=0] at ({\i*\gap+\bw/2},-0.02) {\lab};
  }
  \draw (-0.1,0) -- ({5*\gap+\bw+0.1},0);
\end{scope}

\draw[-{Latex[length=2.2mm]},thick,dashed] (model.east) to[out=25,in=180] (cwm.west);
\draw[-{Latex[length=2.2mm]},thick] (model.east) to[out=-25,in=180]
      ($(ourspanel.west)+(0,0.0)$);
\end{tikzpicture}}
\caption{Concurrent execution has many valid next steps. Prior Code World
Models (dashed) predict one next event and train on it with
cross-entropy. That is ill-defined when the same prefix can produce
several events. We instead target a distribution over next events
(solid), estimated from repeated runs. The illustrative distribution for
a goroutine-leak program has $P(\punblock)=0$, which follows from the
scheduler rather than from learning (Sect.~\ref{sec:signature}).}
\label{fig:teaser}
\end{figure}

\section{Background and Related Work}\label{sec:related}

\paragraph{Execution-aware models of code.}
CodeExecutor pre-trains a Transformer to predict Python execution
traces~\cite{liu2023codeexec}. TRACED adds execution-aware pre-training
objectives~\cite{ding2024traced}. NExT~\cite{ni2024next} and
SemCoder~\cite{ding2024semcoder} train models to reason about execution.
CRUXEval~\cite{gu2024cruxeval} and REval~\cite{chen2025reval} evaluate
output prediction and step-level runtime behaviour. CWM is a 32B model
mid-trained on Python interpreter traces~\cite{cwm}, whose errors
concentrate on long traces and string-valued state~\cite{debugcwm}. All
of these assume one correct trace per input; in our setting the ground
truth is a distribution.

\paragraph{Go concurrency bugs and tools.}
Tu et al.\ studied 171 real Go concurrency bugs and found that message
passing causes as many blocking bugs as shared memory~\cite{tu2019go}.
GoBench turned such bugs into a benchmark that includes the GoKer
kernels used here~\cite{yuan2021gobench}. Static tools detect blocking
misuse of channels~\cite{liu2021gcatch}, behavioural types verify
liveness and safety~\cite{lange2017fencing}. Dynamic tools explore the
schedule space: GFuzz reorders messages~\cite{liu2022gfuzz}, and
Goleak/LeakProf detect goroutine leaks in production
services~\cite{saioc2024goleak}. LLMs have been combined with program
analysis to fix data races at industrial scale~\cite{behrang2025drfix},
and CONCUR benchmarks LLM-generated concurrent code~\cite{concur}. Systematic testing such as CHESS~\cite{musuvathi2008chess} and PCT's
randomised scheduler~\cite{burckhardt2010pct} view a schedule as
something to sample. All of these tools search or repair; none learns a
probabilistic transition model. Testing treats nondeterminism as a search space, whereas we treat
it as a training distribution.

\paragraph{Soft targets and calibration.}
Label smoothing~\cite{muller2019labelsmoothing} and learning from
annotator disagreement~\cite{uma2021disagreement} use distributions rather
than point labels. Language models need fine-tuning to produce diffuse
distributions~\cite{zhang2024diffuse,probcalib}. Calibration is
measured with binned ECE~\cite{naeini2015ece,guo2017calibration}, the
Brier score~\cite{brier1950verification} and NLL. Our contribution is
not distribution matching itself but the \emph{source} of the targets:
the nondeterminism of concurrent execution. In this analogy the
repeated runs act as annotators. Apart from recent preprints on
CWMs~\cite{cwm,debugcwm}, CONCUR~\cite{concur}, distributional
fine-tuning~\cite{probcalib} and the base model~\cite{qwen}, which are
the only descriptions of those systems, all cited work is peer-reviewed.

\section{Methodology}\label{sec:method}

\subsection{Problem Formulation}\label{sec:formulation}

The Go runtime tracer emits scheduler events. We use six event types:
\begin{equation}
\mathcal{E}=\{\pblock,\ \pcreate,\ \pend,\ \psched,\ \pstart,\ \punblock\}.
\end{equation}
A trace $\tau=(s_1,\dots,s_n)$ is a sequence of snapshots, each holding
the event type, the goroutine identifier and the status of every
goroutine (running, runnable, blocked, dead). Given a prefix
$\tau_{1:k}$ and the source $P$, the task is to predict the event type
of $s_{k+1}$. For a group $g$ of runs sharing a program and a prefix
depth, let $c_g(e)$ count next events of type $e$. The target is the
empirical distribution, with a Dirichlet posterior under a Jeffreys
prior ($\alpha=0.5$):
\begin{equation}
\hat{p}_g(e)=\frac{c_g(e)}{\sum_{e'}c_g(e')},\qquad
\tilde{\alpha}_g(e)=\alpha + c_g(e).
\end{equation}
We approximate ``same prefix'' by the same relative depth (25, 50 or
75\% of the trace). The grouped prefixes are therefore comparable but
not byte-identical (Sect.~\ref{sec:threats}).

The formulation has two consequences for evaluation. First, a predictor
should be judged by how well its distribution $q$ matches $\hat p_g$.
We use calibration and the total-variation distance
$\mathrm{TV}(q,\hat p_g)=\tfrac12\sum_e|q(e)-\hat p_g(e)|$. Second, when a test prompt carries labels from several
runs, a deterministic predictor can at best output the mode of those
labels. Top-1 accuracy is therefore bounded by an \emph{oracle ceiling}
(Sect.~\ref{sec:baselines}).

\subsection{Tracing and Empirical Targets}\label{sec:tracing}

Each program is compiled with the race detector and run five times under
the tracer. At each of the three cut points we record the next event.
Runs that deadlock emit no next event and are excluded. The tracer
exposes scheduler events and goroutine states, but not channel buffers,
mutex holders or local variables. A prompt shows the source, the last
(at most ten) events of the prefix as JSON and the current goroutine
states, and asks for the next event in a fixed JSON format. Five runs
was a cost decision, not a statistical one; Sect.~\ref{sec:runs}
measures its consequences.

\subsection{Dataset and Splits}\label{sec:data}

The corpus has 130 Go programs. There are 26 \emph{hand-crafted}
programs (channel, mutex, select, pipeline, wait-group and fan-in/out
patterns, with intentional deadlocks, races and leaks) and 38
\emph{generated} programs from parameterised templates (producer--consumer,
shared lock, pipeline, worker pool, select with timeout) with optional
injected bugs. The remaining 66 are \emph{real-world} GoKer
kernels~\cite{yuan2021gobench} reduced from bugs in CockroachDB,
Kubernetes, gRPC, etcd, Istio, Moby and other systems. After excluding
programs that deadlock at every cut, 63 training programs and 54 GoKer
programs remain. All GoKer programs are held out, so the test set
measures generalisation to unseen real-world code.

Training uses \nTrain{} examples (one per run and cut). The test set has
\nTest{} labels belonging to \uniqPrompts{} unique prompts (54 programs
$\times$ 3 depths); each prompt carries the next events of about five
runs. Test support is \pstart{} 283, \pblock{} 209, \pcreate{} 169,
\psched{} 56, \punblock{} 48 and \pend{} 33. In training, \pcreate{}
(14), \psched{} (4) and \pend{} (17) are rare.

\paragraph{Why so little data?}
Traces are cheap, but programs are not: another run of an existing
program samples the same scheduler distribution, so the effective sample
size is closer to the number of programs and depths than to the number
of examples. We do not claim the data are sufficient
(Sects.~\ref{sec:overfit} and~\ref{sec:threats}).

\subsection{Models and Training Objectives}\label{sec:models}

\paragraph{Model choice.}
We fine-tune \emph{Qwen2.5-Coder-7B-Instruct}~\cite{qwen} for four
reasons:
\begin{enumerate}
\item it has open weights, so the adapters can be released and
  re-evaluated;
\item it is specialised for code, which matters because prompts include
  program source;
\item at 7B it fits 4-bit fine-tuning with a 4{,}096-token context on a
  single 20\,GB GPU (an RTX~4000~Ada at about US\$0.26/h);
\item it follows instructions reliably enough that its JSON output
  parses.
\end{enumerate}
Running the same base zero-shot isolates the effect of fine-tuning.
\emph{Gemini 3.5 Flash}, a strong, low-cost proprietary model, is a
zero-shot reference with and without its thinking mode. It shows whether
a general frontier model already solves the task without training.

\paragraph{Objectives.}
The CE model uses LoRA~\cite{hu2022lora} on a 4-bit base
(QLoRA~\cite{dettmers2023qlora}) with rank 16, $\alpha=32$, learning rate
$2\times10^{-4}$ (cosine schedule), an effective batch size of 8 and
three epochs. The loss is cross-entropy over the response tokens. Each
example is one run's next event, so a group observed five times
contributes labels in proportion to $\hat p_g$. The KL model adds a
distributional term at the token that discriminates between event types:
\begin{equation}
\mathcal{L}=\mathcal{L}_{\mathrm{CE}}+\lambda\,\mathrm{KL}(\hat p_g\,\|\,q),\qquad
q=\mathrm{softmax}(z),
\end{equation}
where $z$ are the logits of the six event-type tokens and $\lambda=1$.
All event names share the leading sub-word ``Go'', so the term is placed
at the second token. The base model, adapters and data are otherwise
identical. Each objective was trained once with one seed.

\paragraph{Multi-step probe.}
Predictions were also fed back for up to 15 steps, and a symbolic
scheduler model counted valid transitions (for example, a blocked
goroutine cannot start) before the first violation.

\section{Experimental Setup and Results}\label{sec:results}

\subsection{Metrics and Statistics}\label{sec:metrics}

We report top-1 accuracy, macro-F1, balanced accuracy (BA, the mean
per-class recall) and per-class recall. The \nTest{} labels are
clustered in 54 programs, so 95\% intervals use a \emph{program-clustered}
bootstrap~\cite{efron1979bootstrap} with 4{,}000 resamples of programs.
For accuracies from the submitted paper, where no per-example
predictions survive, we give Wilson intervals~\cite{wilson1927probable}.
These ignore clustering and are too narrow. Paired comparisons use the
exact McNemar test~\cite{mcnemar1947note}.
Calibration is reported as 10-bin ECE~\cite{naeini2015ece,guo2017calibration},
multi-class Brier score~\cite{brier1950verification}, NLL and mean
confidence. Distributional fit is the mean TV distance to each prompt's
empirical label distribution.

\paragraph{Two evaluations.}
\emph{As originally reported}: the submitted numbers, obtained with
greedy decoding on a bitsandbytes 4-bit base on CUDA. Gemini could not
be re-run and appears only here. \emph{Re-evaluation}: the released CE
and KL adapters~\cite{weave-ce,weave-kl} and the zero-shot base on a
Q4\_K\_M GGUF quantisation with llama.cpp. We force the prefix
\texttt{\{"event\_type": "Go} and read the next-token distribution,
which gives full six-class probabilities. The prediction is the most
probable class. Quantisation and decoding differ from the original run,
so small differences are expected; we report both.

\subsection{Baselines and the Oracle Ceiling}\label{sec:baselines}

The baselines are fitted on the training examples and see only the trace.
\emph{Majority} always predicts \pstart. \emph{Prior sampling} draws
from the training label frequencies; we report its expected accuracy.
\emph{Event n-grams} ($n=1,2,3$) predict the most frequent next event
after the last $n$ event types in the prompt, backing off to shorter
contexts. The \emph{oracle ceiling} predicts each prompt's most frequent
test label. It is not usable, but it is the best top-1 accuracy any
deterministic predictor can reach, because all labels of a prompt share
one input. The ceiling is \oracleCeil\%: even a perfect scheduler model
gets about four in ten labels wrong, because other runs did something
else. Top-1 accuracy is the wrong headline metric here.

\subsection{Main Results}\label{sec:main}

\begin{table}[t]
\centering
\caption{Next-event prediction on the held-out GoKer set. CI: 95\%
program-clustered bootstrap, or Wilson ($^{w}$) where no per-example data
exist. BA: balanced accuracy (\%). $n$: scored test labels; ``--'': not
available.}
\label{tab:main}
\small\setlength{\tabcolsep}{4pt}
\begin{tabular}{lcccc}
\toprule
Predictor & Top-1 (\%) & 95\% CI & Macro-F1 & BA\\
\midrule
\multicolumn{5}{l}{\emph{Baselines (trace only), $n=\nTest$}}\\
Majority (\pstart) & \accMaj & [\loMaj, \hiMaj] & \mfMaj & \baMaj\\
Prior sampling (expected) & \accPrior & -- & \mfPrior & \baPrior\\
Event 1-gram & \accNgram & [\loNgram, \hiNgram] & \mfNgram & \baNgram\\
Event 2-gram & \accNgramTwo & [\loNgramTwo, \hiNgramTwo] & \mfNgramTwo & \baNgramTwo\\
Event 3-gram & \accNgramThree & [\loNgramThree, \hiNgramThree] & \mfNgramThree & \baNgramThree\\
Oracle ceiling (per-prompt mode) & \oracleCeil & -- & -- & --\\
\midrule
\multicolumn{5}{l}{\emph{As originally reported (greedy decoding), $n=\nTest$}}\\
Qwen2.5-Coder-7B zero-shot & 28.6 & [\loRepZS, \hiRepZS]$^{w}$ & -- & --\\
Gemini 3.5 Flash, thinking & 34.8 & [\loRepGemTh, \hiRepGemTh]$^{w}$ & -- & --\\
Gemini 3.5 Flash, no thinking & 35.2 & [\loRepGemNo, \hiRepGemNo]$^{w}$ & -- & --\\
Qwen2.5-Coder-7B KL LoRA & 35.8 & [\loRepKL, \hiRepKL]$^{w}$ & -- & \baRepKL$^{\dagger}$\\
Qwen2.5-Coder-7B CE LoRA & 36.2 & [\loRepCE, \hiRepCE]$^{w}$ & -- & --\\
\midrule
\multicolumn{5}{l}{\emph{Re-evaluation (Q4\_K\_M, llama.cpp, class probabilities)}}\\
Zero-shot ($n=\nZS$) & \accZS & [\loZS, \hiZS] & \mfZS & \baZS\\
CE LoRA ($n=\nCE$) & \accCE & [\loCE, \hiCE] & \mfCE & \baCE\\
KL LoRA ($n=\nKL$) & \accKL & [\loKL, \hiKL] & \mfKL & \baKL\\
\bottomrule
\multicolumn{5}{l}{\footnotesize $^{\dagger}$Mean of the per-event accuracies reported for KL (Table~\ref{tab:perclass}).}
\end{tabular}
\end{table}

\begin{figure}[t]
\centering
\includegraphics[width=\linewidth]{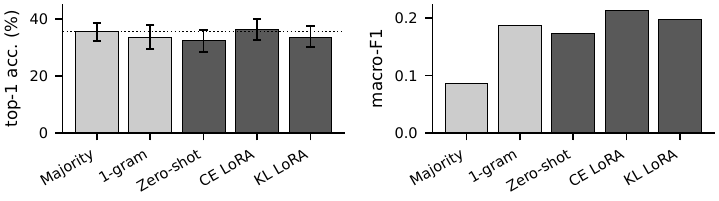}
\caption{Top-1 accuracy with 95\% program-clustered intervals (left) and
macro-F1 (right). Light bars: trace-only baselines. Dark bars:
re-evaluated models. Dotted line: majority rate. The oracle ceiling
(\oracleCeil\%) is not drawn.}
\label{fig:accuracy}
\end{figure}

Table~\ref{tab:main} and Fig.~\ref{fig:accuracy} support three
observations.

\emph{First, the ``ceiling near 35--36\%'' in the submitted paper is the
majority-class rate.} Always predicting \pstart{} scores \accMaj\%
[\loMaj, \hiMaj]. The originally reported CE (36.2\%) and KL (35.8\%)
results and both Gemini configurations lie inside this interval. We
withdraw the submitted claims that fine-tuning ``beats'' Gemini and that
the 7B model generalises better on this metric. The differences among
CE, KL and Gemini are within sampling error, and none is distinguishable
from the majority predictor. The re-evaluation agrees: CE scores
\accCE\%, KL \accKL\% and the untuned base \accZS\%, all within or near
the majority interval. The originally reported gap of the untuned base
(28.6\%) is not reproduced.

\emph{Second, trace statistics alone do about as well.} Prompt-level
n-grams reach \accNgramTwo--\accNgramThree\% without reading the
program.

\emph{Third, metrics that do not reward the majority class are more
informative.} The majority predictor has macro-F1 \mfMaj{} and BA
\baMaj\% by construction, and the n-grams reach a macro-F1 of
\mfNgramTwo--\mfNgram.  Exact McNemar $p$-values in the re-evaluation are
\mcCeVsMajority{} for CE versus majority, \mcCeVsKl{} for CE versus KL,
and \mcCeVsZeroshot{} for CE versus zero-shot. With one seed per
objective, these tests compare two trained adapters, not two training
procedures.

\subsection{Per-Class Results}\label{sec:perclass}

\begin{table}[t]
\centering
\caption{Per-class recall (\%) on the held-out set. ``Rep.\ KL'' is the
per-event accuracy of the KL model as originally reported; ``Train''
is each event's share of the training labels.}
\label{tab:perclass}
\small\setlength{\tabcolsep}{3.2pt}
\begin{tabular}{lrrrrrrr}
\toprule
Event ($n$) & Train & Majority & 1-gram & Zero-shot & CE & KL & Rep.\ KL\\
\midrule
\pstart{} (283)   & 42\%  & \recMajStart & \recNgramStart & \recZSStart & \recCEStart & \recKLStart & 47\\
\pblock{} (209)   & 37\%  & \recMajBlock & \recNgramBlock & \recZSBlock & \recCEBlock & \recKLBlock & 36\\
\pcreate{} (169)  & 1.5\% & \recMajCreate & \recNgramCreate & \recZSCreate & \recCECreate & \recKLCreate & 44\\
\psched{} (56)    & 0.4\% & \recMajSched & \recNgramSched & \recZSSched & \recCESched & \recKLSched & 0\\
\punblock{} (48)  & 17\%  & \recMajUnblock & \recNgramUnblock & \recZSUnblock & \recCEUnblock & \recKLUnblock & 8\\
\pend{} (33)      & 2\%   & \recMajEnd & \recNgramEnd & \recZSEnd & \recCEEnd & \recKLEnd & 0\\
\bottomrule
\end{tabular}
\end{table}

\begin{figure}[t]
\centering
\includegraphics[width=\linewidth]{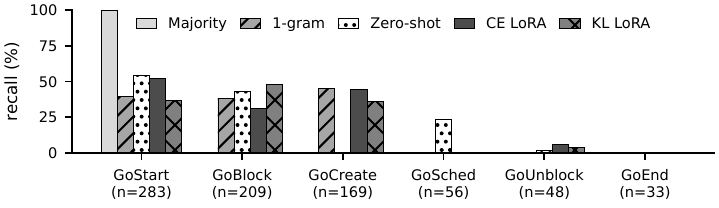}
\caption{Per-class recall for the baselines and the re-evaluated models
(test support $n$ under each event).}
\label{fig:perclass}
\end{figure}

Table~\ref{tab:perclass} and Fig.~\ref{fig:perclass} break the results
down by event type. \emph{Rare events are not learned.} \psched{} and
\pend{} make up 0.4\% and 2\% of training labels. They were never
predicted correctly in the original evaluation (Rep.\ KL column) or by
the trace-only baselines, and \punblock{} was rarely correct. A KL target cannot fix
this, because the empirical training distributions are themselves
dominated by common events. \emph{The test distribution is shifted.}
\pcreate{} (21\%) and \psched{} (7\%) are much more frequent in GoKer
than in training. \emph{The \pcreate{} argument needs qualification.}
The submitted paper took 44\% \pcreate{} accuracy from 1.5\% of training
labels as evidence that the model reads program structure. The last-event
1-gram, however, recovers \recNgramCreate\% of \pcreate{} events from the
trace alone, because programs that spawn workers in a loop emit runs of
\pcreate. We no longer treat \pcreate{} recall as evidence of structural
reasoning beyond what the trace already shows.

\subsection{Calibration}\label{sec:calibration}

\begin{figure}[t]
\centering
\begin{minipage}[b]{0.55\linewidth}\centering
\includegraphics[width=\linewidth]{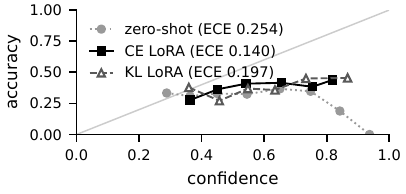}
\end{minipage}\hfill
\begin{minipage}[b]{0.43\linewidth}\centering
\includegraphics[width=\linewidth]{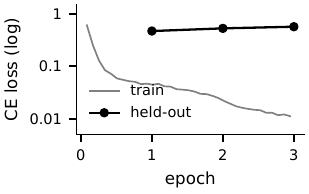}
\end{minipage}
\caption{Left: reliability diagram (10 equal-width confidence bins with
at least five predictions) for the re-evaluated models; each marker is the
accuracy in one bin and the grey diagonal is perfect calibration. Right: CE adapter training log (log scale), with
training loss and loss on held-out GoKer examples logged once per epoch
(Sect.~\ref{sec:overfit}).}
\label{fig:reliability}
\end{figure}

The submitted paper reported an ECE reduction from 0.205 to 0.169 and
attributed it to distribution training. That figure came from a Gemini
prompting study (point versus distribution prompts with a thinking
budget), not from the CE and KL adapters, and it was a mean absolute
probability error, not binned ECE. We have removed it, together with the
entropy--nondeterminism correlation from the same study (Spearman
$\rho=0.412$), which also did not involve the adapters.
Table~\ref{tab:calib} and Fig.~\ref{fig:reliability} (left) replace it with
calibration measured on the adapters and the zero-shot base.

\begin{table}[t]
\centering
\caption{Calibration and distributional fit (re-evaluation). ECE: 10-bin
expected calibration error. Conf.: mean top-class probability (\%). TV:
mean total-variation distance to each prompt's empirical label
distribution. Lower is better, except for Conf.}
\label{tab:calib}
\small
\begin{tabular}{lccccc}
\toprule
Predictor & ECE & Brier & NLL & Conf. & TV\\
\midrule
Training prior (constant) & \ecePrior & \brierPrior & \nllPrior & -- & \tvPrior\\
Event 1-gram distribution & \eceNgram & \brierNgram & \nllNgram & -- & \tvNgram\\
Majority (one-hot) & -- & -- & -- & 100 & \tvMaj\\
Zero-shot & \eceZS & \brierZS & \nllZS & \confZS & \tvZS\\
CE LoRA & \eceCE & \brierCE & \nllCE & \confCE & \tvCE\\
KL LoRA & \eceKL & \brierKL & \nllKL & \confKL & \tvKL\\
\bottomrule
\end{tabular}
\end{table}

The constant prior and the 1-gram's conditional distribution are cheap
references a useful model should match. The comparison relevant to the
original claim is CE against KL. It rests on
one adapter per objective, and ten-bin ECE over \uniqPrompts{} distinct
prompts is itself a noisy statistic. We therefore count a CE--KL
difference as evidence only if ECE, Brier score, NLL and TV agree in
direction. They do: CE is better than KL on all four (ECE \eceCE{}
versus \eceKL, Brier \brierCE{} versus \brierKL, NLL \nllCE{} versus
\nllKL, TV \tvCE{} versus \tvKL), so distribution training did not
improve calibration here. All three models are overconfident (mean
confidence \confCE--\confZS\% against accuracies of \accZS--\accCE\%),
and none beats the constant prior on ECE.

\subsection{Sensitivity to the Number of Runs}\label{sec:runs}

Are five runs enough? We re-ran \sensPrograms{} programs (training
programs and GoKer kernels) 100 times each, with the same tracer, build
flags and cut points. We kept the \sensGroups{} program--depth groups
with at least 60 usable runs (median \sensMedianRuns) and took each
group's distribution over all of its runs as the reference. For each
$K$ we drew $K$ runs 200 times per group and measured the TV distance
to the reference and whether the modes agree.

\begin{table}[t]
\centering
\caption{Distance between a $K$-run empirical target and the reference
from about 100 runs, averaged over \sensGroups{} groups (200 draws
each). The reference has a mean entropy of \refEntropy{} bits, and
\sensDetGroups{} groups are deterministic.}
\label{tab:runs}
\small
\begin{tabular}{rcccc}
\toprule
$K$ & Mean TV & 90th pct.\ TV & Mode agrees (\%) & Entropy (bits)\\
\midrule
1  & \tvKone    & \ptvKone    & \agKone    & \entKone\\
3  & \tvKthree  & \ptvKthree  & \agKthree  & \entKthree\\
5  & \tvKfive   & \ptvKfive   & \agKfive   & \entKfive\\
10 & \tvKten    & \ptvKten    & \agKten    & \entKten\\
20 & \tvKtwenty & \ptvKtwenty & \agKtwenty & \entKtwenty\\
50 & \tvKfifty  & \ptvKfifty  & \agKfifty  & \entKfifty\\
\bottomrule
\end{tabular}
\end{table}

\begin{figure}[t]
\centering
\includegraphics[width=\linewidth]{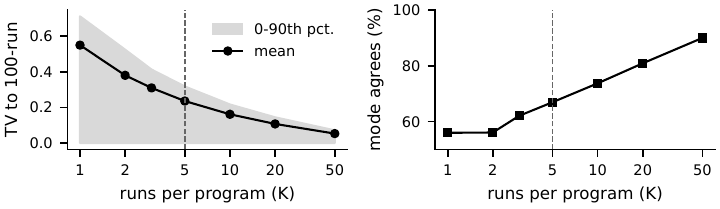}
\caption{Run-count sensitivity. Left: TV distance to the ~100-run
reference (line: mean; band: 0--90th percentile). Right: fraction of
draws whose mode matches the reference mode (axis starts at 50\%). Dashed
line: $K=5$, as used in the dataset.}
\label{fig:runs}
\end{figure}

Table~\ref{tab:runs} and Fig.~\ref{fig:runs} show three things. First,
nondeterminism is substantial: only \sensDetGroups{} of \sensGroups{}
groups are deterministic over about 100 runs, and the mean reference entropy is \refEntropy{} bits. Second, five-run
targets are noisy. Their mean TV distance is \tvKfive, and their mode
matches the reference mode in only \agKfive\% of draws. They also have
lower entropy (\entKfive{} versus \refEntropy{} bits), so they are biased
towards sparse distributions in which possible but rare events go
unobserved. Third, the error shrinks as sampling theory predicts: the
mean TV is \tvKtwenty{} at 20 runs and \tvKfifty{} at 50, and mode
agreement rises to \agKtwenty\% and \agKfifty\%. Two disjoint halves of
a reference sample differ by a mean TV of \splitHalfTV.

Five runs are therefore a cheap first estimate, not an adequate target.
A KL objective trained on targets about \tvKfive{} TV away from what it
should learn has a weak signal to exploit, which may explain why the
KL objective showed no clear advantage over CE, either as originally
reported (35.8\% versus 36.2\%) or in the re-evaluation (\accKL\% versus
\accCE\%, McNemar $p=\mcCeVsKl$). We recommend at least 20 runs per group.

\subsection{Overfitting}\label{sec:overfit}

Figure~\ref{fig:reliability} (right) plots the trainer log released
with the CE adapter. Training loss falls from 0.60 to 0.01 over three epochs (357
steps). Held-out loss rises after every epoch: 0.468, 0.527 and 0.566.
The model memorises the training set after the first epoch. This loss
was only monitored: the final epoch-3 adapter is the one evaluated, so
the test set played no part in model selection. Without a validation
split drawn from the training programs, however, early stopping was not
possible. The KL run logged no held-out loss; its training loss, which
is dominated by the KL term, falls from 15.5 to 1.6.

\section{Discussion}\label{sec:discussion}

\subsection{Where the Ceiling Comes From}\label{sec:ceiling}

The cluster of results near 35--36\% that the submitted paper read as a
finding about task difficulty has four simpler causes, each with a
remedy: (i) the metric, since top-1 accuracy rewards the dominant
\pstart{} class and is capped at \oracleCeil\% by nondeterminism
(remedy: distributional metrics); (ii) class imbalance and shift, since
rare training events are rarely if ever predicted (rebalanced or targeted
programs); (iii) noisy five-run targets with a TV error of \tvKfive{}
(20 or more runs per group); and (iv) small, memorised training data
(more programs and a validation split).

\subsection{The Select-Block Leak Signature}\label{sec:signature}

One pattern in the data has a formal explanation. Consider a goroutine
$G$ that enters a \texttt{select} at time $t$ where no case can be
satisfied by the remaining execution; we call this a \emph{select-block
leak}. A \punblock{} for $G$ fires only when another goroutine sends on
or closes a channel $G$ waits on, or releases a lock $G$ waits for. If no
reachable goroutine can do so, no \punblock{} occurs for $G$ after $t$.
When no other goroutine can be unblocked either, $P(\punblock)=0$ for
every cut after $G$'s first \pblock. This holds regardless of the number
of \texttt{select} cases (we observed it for two-case and four-case
selects), and it agrees with static reasoning about channel
constraints~\cite{liu2021gcatch,lange2017fencing}. In the data, all five
runs of such programs show $P(\punblock)=0$ at every depth. The
signature is mechanism-specific. Leaks that do useful work before
blocking show $P(\punblock)>0$ at early depths. Used as an unsupervised
detector, the signal separated buggy from clean programs only weakly
(Cohen's $d=0.29$, $n=9$, not significant; as originally reported). It
characterises one leak class exactly and gives a sanity check for
learned models, but it is not a detector.

\subsection{Multi-Step Coherence}\label{sec:rollout}

\begin{table}[t]
\centering
\caption{Multi-step probe on GoKer (15 steps, 3 samples per program), as
originally reported for the KL model. Survival is the mean number of
valid transitions before the first violation. No per-rollout data were
retained, so no intervals are available.}
\label{tab:rollout}
\small
\begin{tabular}{lcccc}
\toprule
Outcome & $n$ & Dominant event & Survival (steps) & Entropy (bits)\\
\midrule
Leak & 37 & \pblock{} (50\%) & 1.11 & 0.945\\
Race & 17 & \pblock{} (65\%) & 0.67 & 0.773\\
\bottomrule
\end{tabular}
\end{table}

Table~\ref{tab:rollout} repeats the probe from the submitted paper. On
leak programs the first predicted step is dominated by \pblock{} with
$P(\punblock)$ near zero, as the signature predicts. Mean survival,
however, is about one step: the models are single-step predictors, not
simulators. Per-rollout data were not retained, so no survival curve can
be drawn.

\subsection{Threats to Validity}\label{sec:threats}

\emph{Single seed.} Each objective was trained once, so training
variance could be as large as the CE--KL difference.
\emph{Two evaluations.} The re-evaluation differs in quantisation and
decoding (Sect.~\ref{sec:metrics}). Gemini could not be re-run, and its
Wilson intervals ignore clustering.
\emph{Noisy, approximate targets.} Five-run targets carry a TV error of
about \tvKfive. Grouping by relative depth merges similar but not
identical prefixes, and the labels of a test prompt come from runs whose
prefixes differ slightly from the one shown.
\emph{Partial observability.} Channel buffers, mutex ownership and local
variables are not traced, which limits events such as \punblock.
\emph{Small data and no learning curve.} The data come from 63 training
and 54 test programs. We monitored overfitting but did not measure
accuracy against training-set size. That needs repeated GPU fine-tuning
and is the most important missing experiment.
\emph{External validity.} GoKer kernels are reduced reproductions, not
full applications.

\section{Conclusion}\label{sec:conclusion}

We formulated next-event prediction for concurrent Go programs as
estimation of an empirical distribution over scheduler events, released
a benchmark of real bug kernels, and evaluated CE and KL fine-tuning of a
7B code model against trivial baselines. The result is
neutral-to-negative. On top-1 accuracy the fine-tuned models, both as
originally reported and as re-evaluated, are indistinguishable from
predicting the majority event, trace-only n-grams
come close, and nondeterminism itself caps top-1 accuracy at
\oracleCeil\%. A 100-run study shows that five-run targets are too noisy
to train a distribution-matching objective well. What remains are the
formulation, the benchmark with baselines and an oracle ceiling, the
run-count analysis, and the leak signature. Future work should use 20 or
more runs per group, evaluate with distributional metrics, add and
rebalance programs with a validation split, and trace channel and mutex
state. Code, data and both adapters are
public~\cite{weave-repo,weave-bench,weave-ce,weave-kl}. Fine-tuning cost
about US\$12, and Gemini inference cost about US\$60.

\subsubsection*{Acknowledgments.}
The author thanks the anonymous reviewers for their detailed comments,
which substantially changed the evaluation in this paper.

\subsubsection*{Disclosure of Interests.}
The author has no competing interests to declare that are relevant to the
content of this article.

\bibliographystyle{splncs04}
\bibliography{references}

\end{document}